\documentclass{article}
\usepackage{spconf,amsmath,graphicx}
\usepackage{multirow}

\title{Deep LSTM for Large Vocabulary Continuous Speech Recognition}
\name{Xu Tian, Jun Zhang, Zejun Ma, Yi He, Juan Wei, Peihao Wu, Wenchang Situ, Shuai Li, Yang Zhang}

\address{Alibaba Shenma Search, Beijing, China\\
	\small \texttt\{xu.tian, zj102217, zejun.mamzj, heyi.hy, wj80290, peihao.wph, wenchang.situwc, voolc.li, zy80232\}@alibaba-inc.com}

\begin{document}

\maketitle

\begin{abstract}

Recurrent neural networks (RNNs), especially long short-term memory (LSTM) RNNs, are effective network for sequential task like speech recognition. Deeper LSTM models perform well on large vocabulary continuous speech recognition, because of their impressive learning ability. However, it is more difficult to train a deeper network. We introduce a training framework with layer-wise training and exponential moving average methods for deeper LSTM models. 
It is a competitive framework that LSTM models of more than 7 layers are successfully trained on Shenma voice search data in Mandarin and they outperform the deep LSTM models trained by conventional approach.
Moreover, in order for online streaming speech recognition applications, the shallow model with low real time factor is distilled from the very deep model. The recognition accuracy have little loss in the distillation process. 
Therefore, the model trained with the proposed training framework reduces relative 14\% character error rate, compared to original model which has the similar real-time capability. Furthermore, the novel transfer learning strategy with segmental Minimum Bayes-Risk is also introduced in the framework. The strategy makes it possible that 
training with only a small part of dataset could outperform full dataset training from the beginning.

\end{abstract}

\section{Introduction}
\label{sec:intro}

Recently, deep neural network has been widely employed in various recognition tasks. Increasing the depth of neural network is a effective way to improve the performance, and convolutional neural network (CNN) has benefited from it in visual recognition task\cite{he2015deep}. Deeper long short-term memory (LSTM) recurrent neural networks (RNNs) are also applied in large vocabulary continuous speech recognition (LVCSR) task, because LSTM networks have shown better performance than Fully-connected feed-forward deep neural network\cite{hinton2012deep,graves2013speech,graves2013hybrid,sak2014long}.

Training neural network becomes more challenge when it goes deep. A conceptual tool called linear classifier probe is introduced to better understand the dynamics inside a neural network \cite{alain2016understanding}. The discriminating features of linear classifier is the hidden units of a intermediate layer. For deep neural networks, it is observed that deeper layer's accuracy is lower than that of shallower layers. Therefore, the tool shows the difficulty of deep neural model training visually.

Layer-wise pre-training is a successful method to train very deep neural networks \cite{hinton2006reducing}. The convergence becomes harder with increasing the number of layers, even though the model is initialized with Xavier or its variants \cite{glorot2010understanding,he2015delving}. But the deeper network which is initialized with a shallower trained network could converge well.

The size of LVCSR training dataset goes larger and training with only one GPU becomes high time consumption inevitably. Therefore, parallel training with multi-GPUs is more suitable for LVCSR system. Mini-batch based stochastic gradient descent (SGD) is the most popular method in neural network training procedure. Asynchronous SGD is a successful effort for parallel training based on it \cite{dean2012large,zhang2013asynchronous}. It can many times speed up the training time without decreasing the accuracy. Besides, synchronous SGD is another effective effort, where the parameter server waits for every works to finish their computation and sent their local models to it, and then it sends updated model back to all workers \cite{chen2016revisiting}. Synchronous SGD converges well in parallel training with data parallelism, and is also easy to implement.

In order to further improve the performance of deep neural network with parallel training, several methods are proposed. Model averaging method achieves linear speedup, as the final model is averaged from all parameters of local models in different workers \cite{mcdonald2010distributed,zinkevich2010parallelized}, but the accuracy decreases compared with single GPU training. Moreover, blockwise model-updating filter (BMUF) provides another almost linear speedup approach with multi-GPUs on the basis of model averaging. It can achieve improvement or no-degradation of recognition performance compared with mini-batch SGD on single GPU \cite{chen2016scalable}.

Moving averaged (MA) approaches are also proposed for parallel training. It is demonstrated that the moving average of the parameters obtained by SGD performs as well as the parameters that minimize the empirical cost, and moving average parameters can be used as the estimator of them, if the size of training data is large enough \cite{polyak1992acceleration}. One pass learning is then proposed, which is the combination of learning rate schedule and averaged SGD using moving average \cite{xu2011towards}. Exponential moving average (EMA) is proposed as a non-interference method\cite{tian2017}. EMA model is not broadcasted to workers to update their local models, and it is applied as the final model of entire training process. EMA method is utilized with model averaging and BMUF to further decrease the character error rate (CER). It is also easy to implement in existing parallel training systems.

Frame stacking can also speed up the training time \cite{sak2015fast}. The super frame is stacked by several regular frames, and it contains the information of them. Thus, the network can see multiple frames at a time, as the super frame is new input. Frame stacking can also lead to faster decoding.

For streaming voice search service, it needs to display intermediate recognition results while users are still speaking.
As a result, the system needs to fulfill high real-time requirement, and we prefer unidirectional LSTM network rather than bidirectional one.
High real-time requirement means low real time factor (RTF), but the RTF of deep LSTM model is higher inevitably. The dilemma of recognition accuracy and real-time requirement is an obstacle to the employment of deep LSTM network. Deep model outperforms because it contains more knowledge, but it is also cumbersome. As a result, the knowledge of deep model can be distilled to a shallow model \cite{hinton2015distilling}. It provided a effective way to employ the deep model to the real-time system.

In this paper, we explore a entire deep LSTM RNN training framework, and employ it to real-time application. The deep learning systems benefit highly from a large quantity of labeled training data. Our first and basic speech recognition system is trained on 17000 hours of Shenma voice search dataset. It is a generic dataset sampled from diverse aspects of search queries. The requirement of speech recognition system also addressed by specific scenario, such as map and navigation task. The labeled dataset is too expensive, and training a new model with new large dataset from the beginning costs lots of time. Thus, it is natural to think of transferring the knowledge from basic model to new scenario's model. Transfer learning expends less data and less training time than full training. In this paper, we also introduce a novel transfer learning strategy with segmental Minimum Bayes-Risk (sMBR). As a result, transfer training with only 1000 hours data can match equivalent performance for full training with 7300 hours data. 

Our deep LSTM training framework for LVCSR is presented in Section 2. Section 3 describes how the very deep models does apply in real world applications, and how to transfer the model to another task. The framework is analyzed and discussed in Section 4, and followed by the conclusion in Section 5.

\section{Our Training Framework}
\subsection{Layer-wise Training with Soft Target and Hard Target}
Gradient-based optimization of deep LSTM network with random initialization get stuck in poor solution easily. Xavier initialization can partially solve this problem \cite{glorot2010understanding}, so this method is the regular initialization method of all training procedure. However, it does not work well when it is utilized to initialize very deep model directly, because of vanishing or exploding gradients. Instead, layer-wise pre-training method is a effective way to train the weights of very deep architecture\cite{hinton2006reducing,hinton2006fast}. 
In layer-wise pre-training procedure, a one-layer LSTM model is firstly trained with normalized initialization. Sequentially, two-layers LSTM model's first layer is initialized by trained one-layer model, and its second layer is regularly initialized. In this way, a deep architecture is layer-by-layer trained, and it can converge well. 

In conventional layer-wise pre-training, only parameters of shallower network are transfered to deeper one, and the learning targets are still the alignments generated by HMM-GMM system. The targets are vectors that only one state's probability is one, and the others' are zeros. They are known as hard targets, and they carry limited knowledge as only one state is active. In contrast, the knowledge of shallower network should be also transfered to deeper one. It is obtained by the softmax layer of existing model typically, so each state has a probability rather than only zero or one, and called as soft target. As a result, the deeper network which is student network learns the parameters and knowledge from shallower one which is called teacher network. 
When training the student network from the teacher network, the final alignment is the combination of hard target and soft target in our layer-wise training phase. The final alignment provides various knowledge which transfered from teacher network and extracted from true labels. If only soft target is learned, student network perform no better than teacher network, but it could outperform teacher network as it also learns true labels.

The deeper network spends less time to getting the same level of original network than the network trained from the beginning, as a period of low performance is skipped. Therefore, training with hard and soft target is a time saving method. For large training dataset, 
training with the whole dataset still spends too much time. A network firstly trained with only a small part of dataset could go deeper as well, and so the training time reducing rapidly. When the network is deep enough, it then trained on the entire dataset to get further improvement. There is no gap of accuracy between these two approaches, but latter one saves much time.

\subsection{Differential Saturation Check}
The objects of conventional saturation check are gradients and the cell activations \cite{sak2014long}. Gradients are clipped to range [-5, 5], while the cell activations clipped to range [-50, 50]. Apart from them, the differentials of recurrent layers is also limited. If the differentials go beyond the range, corresponding back propagation is skipped, while if the gradients and cell activations go beyond the bound, values are set as the boundary values. The differentials which are too large or too small will lead to the gradients easily vanishing, and it demonstrates the failure of this propagation. As a result, the parameters are not updated, and next propagation .

\subsection{Sequence Discriminative Training}
Cross-entropy (CE) is widely used in speech recognition training system as a frame-wise discriminative training criterion. However, it is not well suited to speech recognition, because speech recognition training is a sequential learning problem. In contrast, sequence discriminative training criterion has shown to further improve performance of neural network first trained with cross-entropy \cite{kingsbury2009lattice,sak2015learning,sak2015acoustic}. We choose state-level minimum bayes risk (sMBR)\cite{kingsbury2009lattice} among a number of sequence discriminative criterion is proposed, such as maximum mutual information (MMI) \cite{normandin1991hidden} and minimum phone error (MPE) \cite{povey2005discriminative}. MPE and sMBR are designed to minimize the expected error of different granularity of labels, while CE aims to minimizes expected frame error, and MMI aims to minimizes expected sentence error. State-level information is focused on by sMBR.

a frame-level accurate model is firstly trained by CE loss function, and then sMBR loss function is utilized for further training to get sequence-level accuracy. Only a part of training dataset is needed in sMBR training phase on the basis of whole dataset CE training.

\subsection{Parallel Training}
It is demonstrated that training with larger dataset can improve recognition accuracy. However, larger dataset means more training samples and more model parameters. Therefore, parallel training with multiple GPUs is essential, and it makes use of data parallelism \cite{dean2012large}. The entire training data is partitioned into several split without overlapping and they are distributed to different GPUs. Each GPU trains with one split of training dataset locally. All GPUs synchronize their local models with model average method after a mini-batch optimization \cite{mcdonald2010distributed,zinkevich2010parallelized}.

\subsubsection{Block-wise Model Updating Filter}
Model average method achieves linear speedup in training phase, but the recognition accuracy decreases compared with single GPU training. Block-wise model updating filter (BMUF) is another successful effort in parallel training with linear speedup as well. It can achieve no-degradation of recognition accuracy with multi-GPUs \cite{chen2016scalable}. In the model average method, aggregated model $\bar \theta(t)$ is computed and broadcasted to GPUs. On the basis of it, BMUF proposed a novel model updating strategy:
$$\bar\theta(t)=\frac{1}{N}\sum^N_{i=1}\theta_i$$
$$G(t)= \bar \theta(t) - \theta_g(t-1)​$$
$$\Delta(t)=\eta_t\Delta(t-1)+\zeta_tG(t)$$
Where $G(t)$ denotes model update, and $\Delta(t)$ is the global-model update. There are two parameters in BMUF, block momentum $\eta$, and block learning rate $\zeta$. Then, the global model is updated as 
$$\theta_g(t)=\theta_g(t-1)+\Delta(t)$$
Consequently, $\theta_g(t)$ is broadcasted to all GPUs to initial their local models, instead of $\bar \theta(t)$ in model average method.

\subsubsection{Exponential Moving Average Model}
Averaged SGD is proposed to further accelerate the convergence speed of SGD. Averaged SGD leverages the moving average (MA) $\bar\theta$ as the estimator of $\theta^*$ \cite{polyak1992acceleration}:
$$\bar{\theta_t} = \frac{1}{t}\sum_{\tau=1}^{t}\theta_\tau$$ 
Where $\theta_{\tau}$ is computed by model averaging or BMUF. It is shown that $\bar\theta_t$ can well converge to $\theta^*$, with the large enough training dataset in single GPU training. It can be considered as a non-interference strategy that $\bar\theta_t$ does not participate the main optimization process, and only takes effect after the end of entire optimization. 
However, for the parallel training implementation, each $\theta_\tau$ is computed by model averaging and BMUF with multiple models, and moving average model $\bar\theta_t$ does not well converge, compared with single GPU training.

Model averaging based methods are employed in parallel training of large scale dataset, because of their faster convergence, and especially no-degradation implementation of BMUF. But combination of model averaged based methods and moving average does not match the expectation of further enhance performance and it is presented as
$$\bar\theta_{g_t} = \frac{1}{t}\sum_{\tau=1}^{t}\theta_{g_\tau}$$ 
The weight of each $\theta_{g_t}$ is equal in moving average method regardless the effect of temporal order. But $t$ closer to the end of training achieve higher accuracy in the model averaging based approach, and thus it should be with more proportion in final $\bar \theta_g$. As a result, exponential moving average(EMA) is appropriate, which the weight for each older parameters decrease exponentially, and never reaching zero. 
After moving average based methods, the EMA parameters are updated recursively as
$$\hat \theta_{g_t}=\alpha\hat\theta_{g_{t-1}}+(1-\alpha)\theta_{g_t}$$
Here $\alpha$ represents the degree of weight decrease, and called exponential updating rate. 
EMA is also a non-interference training strategy that is implemented easily, as the updated model is not broadcasted. Therefore, there is no need to add extra learning rate updating approach, as it can be appended to existing training procedure directly.

\section{Deployment}
There is a high real time requirement in real world application, especially in online voice search system. Shenma voice search is one of the most popular mobile search engines in China, and it is a streaming service that intermediate recognition results displayed while users are still speaking. Unidirectional LSTM network is applied, rather than bidirectional one, because it is well suited to real-time streaming speech recognition.

\subsection{Distillation}
It is demonstrated that deep neural network architecture can achieve improvement in LVCSR. However, it also leads to much more computation and higher RTF, so that the recognition result can not be displayed in real time. It should be noted that deeper neural network contains more knowledge, but it is also cumbersome. the knowledge is key to improve the performance. If it can be transfered from cumbersome model to a small model, the recognition ability can also be transfered to the small model. Knowledge transferring to small model is called distillation \cite{hinton2015distilling}. The small model can perform as well as cumbersome model, after distilling. It provide a way to utilize high-performance but high RTF model in real time system. The class probability produced by the cumbersome model is regarded as soft target, and the generalization ability of cumbersome model is transfered to small model with it. Distillation is model's knowledge transferring approach, so there is no need to use the hard target, which is different with the layer-wise training method.

\subsection{Transfer Learning with sMBR}
For a certain specific scenario, the model trained with the data recorded from it has better adaptation than the model trained with generic scenario. But it spends too much time training a model from the beginning, if there is a well-trained model for generic scenarios. Moreover, labeling a large quantity of training data in new scenario is both costly and time consuming. If a model transfer trained with smaller dataset can obtained the similar recognition accuracy compared with the model directly trained with larger dataset, it is no doubt that transfer learning is more practical. Since specific scenario is a subset of generic scenario, some knowledge can be shared between them. Besides, generic scenario consists of various conditions, so its model has greater robustness. 
As a result, not only shared knowledge but also robustness can be transfered from the model of generic scenario to the model of specific one.

As the model well trained from generic scenario achieves good performance in frame level classification, sequence discriminative training is required to adapt new model to specific scenario additionally. Moreover, it does not need alignment from HMM-GMM system, and it also saves amount of time to prepare alignment.

\section{Experiments}

\subsection{Training Data}
A large quantity of labeled data is needed for training a more accurate acoustic model. We collect the 17000 hours labeled data from Shenma voice search, which is one of the most popular mobile search engines in China. The dataset is created from anonymous online users' search queries in Mandarin, and all audio file's sampling rate is 16kHz, recorded by mobile phones. This dataset consists of many different conditions, such as diverse noise even low signal-to-noise, babble, dialects, accents, hesitation and so on. 

In the Amap, which is one of the most popular web mapping and navigation services in China, users can search locations and navigate to locations they want though voice search.
To present the performance of transfer learning with sequence discriminative training, the model trained from Shenma voice search which is greneric scenario transfer its knowledge to the model of Amap voice search. 7300 hours labeled data is collected in the similar way of Shenma voice search data collection.

Two dataset is divided into training set, validation set and test set separately, and the quantity of them is shown in Table~\ref{table:data}. The three sets are split according to speakers, in order to avoid utterances of same speaker appearing in three sets simultaneously. The test sets of Shenma and Amap voice search are called Shenma Test and Amap Test.

\begin{table}[t]
	\centering	
	\begin{tabular}{|c|c|c|}
		\hline
		Dataset & Shenma Voice Search & Amap  \\ 
		\hline
		\hline
		Training set  & 16150 & 6935 \\
		\hline
		Validation set & 850 & 365  \\
		\hline
		Test set &  10 & 8 \\
		\hline
		\bf Total & 17010 & 7308 \\
		\hline
	\end{tabular}	
	\caption{ \label{table:data} \small The time summation of different sets of Shenma voice search and Amap.}
\end{table}


\subsection{Experimental setup}
LSTM RNNs outperform conventional RNNs for speech recognition system, especially deep LSTM RNNs, because of its long-range dependencies more accurately for temporal sequence conditions \cite{hermans2013training,sak2015acoustic}. Shenma and Amap voice search is a streaming service that intermediate recognition results displayed while users are still speaking. So as for online recognition in real time, we prefer unidirectional LSTM model rather than bidirectional one. Thus, the training system is unidirectional LSTM-based.

A 26-dimensional filter bank and 2-dimensional pitch feature is extracted for each frame, and is concatenated with first and second order difference as the final input of the network. The super frame are stacked by 3 frames without overlapping.
The architecture we trained consists of two LSTM layers with sigmoid activation function, followed by a full-connection layer. The out layer is a softmax layer with 11088 hidden markov model (HMM) tied-states as output classes, the loss function is cross-entropy (CE). The performance metric of the system in Mandarin is reported with character error rate (CER). The alignment of frame-level ground truth is obtained by GMM-HMM system. Mini-batched SGD is utilized with momentum trick and the network is trained for a total of 4 epochs. The block learning rate and block momentum of BMUF are set as 1 and 0.9. 5-gram language model is leveraged in decoder, and the vocabulary size is as large as 760000. Differentials of recurrent layers is limited to range [-10000,10000], while gradients are clipped to range [-5, 5] and cell activations clipped to range [-50, 50]. After training with CE loss, sMBR loss is employed to further improve the performance.

It has shown that BMUF outperforms traditional model averaging method, and it is utilized at the synchronization phase. After synchronizing with BMUF, EMA method further updates the model in non-interference way. The training system is deployed on the MPI-based HPC cluster where 8 GPUs. Each GPU processes non-overlap subset split from the entire large scale dataset in parallel. 

\begin{figure}[ht]
	\vskip 0.2in
	\begin{center}
		\includegraphics[width=\columnwidth]{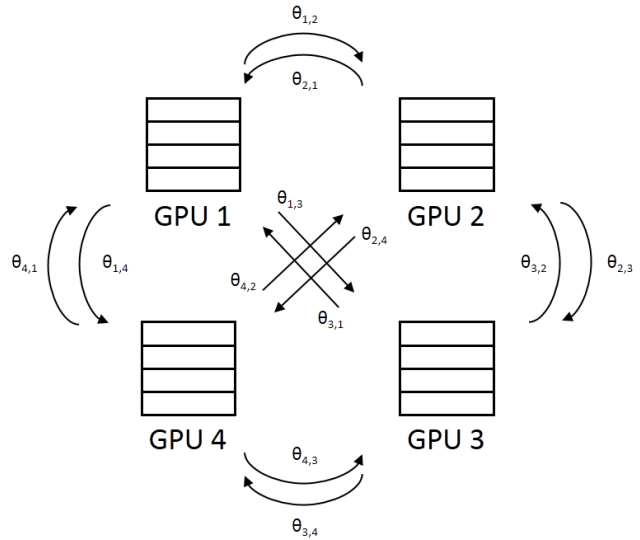}
		\caption{Mesh AllReduce of model averaging and BMUF.}
		\label{fig:reduce}
	\end{center}
	\vskip -0.2in
\end{figure}

Local models from distributed workers synchronize with each other in decentralized way. In the traditional model averaging and BMUF method, a parameter server waits for all workers to send their local models, aggregate them, and send the updated model to all workers. Computing resource of workers is wasted until aggregation of the parameter server done. Decentralized method makes full use of computing resource, and we employ the MPI-based Mesh AllReduce method. It is mesh topology as shown in Figure~\ref{fig:reduce}. There is no centralized parameter server, and peer to peer communication is used to transmit local models between workers. Local model $\theta_i$ of $i$-th worker in $N$ workers cluster is split to $N$ pieces $\theta_{i,j}\ j=1\cdots N$, and send to corresponding worker. In the aggregation phase, $j$-th worker computed $N$ splits of model $\theta_{i,j}\ i=1\cdots N$ and send updated model $\bar \theta_{g_j}$ back to workers. As a result, all workers participate in aggregation and no computing resource is dissipated. It is significant to promote training efficiency, when the size of neural network model is too large. The EMA model is also updated additionally, but not broadcasting it.

\section{Results}
In order to evaluate our system, several sets of experiments are performed. The Shenma test set including about 9000 samples and Amap test set including about 7000 samples contain various real world conditions. It simulates the majority of user scenarios, and can well evaluates the performance of a trained model.
Firstly, we show the results of models trained with EMA method. Secondly, for real world applications, very deep LSTM is distilled to a shallow one, so as for lower RTF. The model of Amap is also needed to train for map and navigation scenarios. The performance of transfer learning from Shenma voice search to Amap voice search is also presented.

\subsection{Layer-wise Training}
In layer-wise training, the deeper model learns both parameters and knowledge from the shallower model. The deeper model is initialized by the shallower one, and its alignment is the combination of hard target and soft target of shallower one. Two targets have the same weights in our framework. The teacher model is trained with CE. For each layer-wise trained CE model, corresponding sMBR model is also trained, as sMBR could achieve additional improvement. In our framework, 1000 hours data is randomly selected from the total dataset for sMBR training. There is no obvious performance enhancement when the size of sMBR training dataset increases.

For very deep unidirectional LSTM initialized with Xavier initialization algorithm, 6-layers model converges well, but there is no further improvement with increasing the number of layers. Therefore, the first 6 layers of 7-layers model is initialized by 6-layers model, and soft target is provided by 6-layers model. Consequently, deeper LSTM is also trained in the same way. It should be noticed that the teacher model of 9-layers model is the 8-layers model trained by sMBR, while the other teacher model is CE model. As shown in Table~\ref{table:smbr}, the layer-wise trained models always performs better than the models with Xavier initialization, as the model is deep.
Therefore, for the last layer training, we choose 8-layers sMBR model as the teacher model instead of CE model.
A comparison between 6-layers and 9-layers sMBR models shows that 3 additional layers of layer-wise training brings relative 12.6\% decreasing of CER.
It is also significant that the averaged CER of sMBR models with different layers decreases absolute 0.73\% approximately compared with CE models, so the improvement of sequence discriminative learning is promising.

\subsection{Distillation}

\begin{table}[t]
	\centering	
	\noindent
	\renewcommand{\multirowsetup}{\centering}  
	\begin{tabular}{|c|c|c|c|}  
		\hline
		\multirow{2}{0.7cm}{\textbf{Layer}} & \multicolumn{3}{c}{\textbf{CER (\%)}} \\ \cline{2-4}
		& \textbf{Xavier Init CE} & \textbf{Layer-wise CE} & \textbf{CE+sMBR} \\ 
		\hline
		\hline
		6 & 3.72  & - & 2.85 \\
		\hline
		7 & 3.93 & 3.68 & 2.81  \\
		\hline
		8 & 3.81 & 3.60 & 2.77 \\
		\hline
		9 & 3.87 & 2.82 &  2.49 \\
		\hline
	\end{tabular}
	\caption{ \label{table:smbr} \small The CER of 6 to 9-layers models trained by regular Xavier Initialization, layer-wise training with CE criterion and CE + sMBR criteria. The teacher of 9-layer model is 8-layers sMBR model, while the others' teacher is CE model.}
\end{table}

9-layers unidirectional LSTM model achieves outstanding performance, but meanwhile it is too computationally expensive to allow deployment in real time recognition system. In order to ensure real-time of the system, the number of layers needs to be reduced. The shallower network can learn the knowledge of deeper network with distillation. It is found that RTF of 2-layers network is acceptable, so the knowledge is distilled from 9-layers well-trained model to 2-layers model. Table~\ref{table:distill} shows that distillation from 9-layers to 2-layers brings RTF decrease of relative 53\%, while CER only increases 5\%. The knowledge of deep network is almost transfered with distillation, 
Distillation brings promising RTF reduction, but only little knowledge of deep network is lost. Moreover, CER of 2-layers distilled LSTM decreases relative 14\%, compared with 2-layers regular-trained LSTM.

\begin{table}[t]
	\centering	
	\begin{tabular}{|c|c|c|}
		\hline
		\textbf{Models} & \textbf{CER (\%)} & \textbf{RTF} \\ 
		\hline
		\hline
		9-layers LSTM  & 2.49 & 0.74 \\
		\hline
		2-layers regular-trained LSTM & 3.06 & 0.36  \\
		\hline
		2-laryers distilled LSTM &  2.63 & 0.35 \\
		\hline
	\end{tabular}	
	\caption{ \label{table:distill} \small The CER and RTF of 9-layers, 2-layers regular-trained and 2-laryers distilled LSTM. }
\end{table}

\subsection{Transfer Learning}
2-layers distilled model of Shenma voice search has shown a impressive performance on Shenma Test, and we call it Shenma model. It is trained for generic search scenario, but it has less adaptation for specific scenario like Amap voice search. 
Training with very large dataset using CE loss is regarded as improvement of frame level recognition accuracy, and sMBR with less dataset further improves accuracy as sequence discriminative training. If robust model of generic scenario is trained, there is no need to train a model with very large dataset, and sequence discriminative training with less dataset is enough. Therefore, on the basis of Shenma model, it is sufficient to train a new Amap model with small dataset using sMBR. As shown in Table~\ref{table:tranfer}, Shenma model presents the worst performance among three methods, since it does not trained for Amap scenario. 2-layers Shenma model further trained with sMBR achieves about 8.1\% relative reduction, compared with 2-layers regular-trained Amap model. Both training sMBR datasets contain the same 1000 hours data. As a result, with the Shenma model, only about 14\% data usage achieves lower CER, and it leads to great time and cost saving with less labeled data. Besides, transfer learning with sMBR does not use the alignment from the HMM-GMM system, so it also saves huge amount of time.

\begin{table}[t]
	\centering	
	\begin{tabular}{|c|c|}
		\hline
		Training methods & CER (\%) \\ 
		\hline
		\hline
		Shenma model & 7.87 \\
		\hline
		Amap CE + sMBR & 6.81   \\
		\hline
		Shenma model + Amap sMBR & 6.26  \\
		\hline
	\end{tabular}	
	\caption{ \label{table:tranfer} \small The CER of different 2-layers models, which are Shenma distilled model, Amap model further trained with Amap dataset, and Shenma model trained with sMBR on Amap dataset. }
\end{table}

\section{Conclusion}
We have presented a whole deep unidirectional LSTM parallel training system for LVCSR. The recognition performance improves when the network goes deep. Distillation makes it possible that deep LSTM model transfer its knowledge to shallow model with little loss. The model could be distilled to 2-layers model with very low RTF, so that it can display the immediate recognition results. As a result, its CER decrease relatively 14\%, compared with the 2-layers regular trained model. In addition, transfer learning with sMBR is also proposed. If a great model has well trained from generic scenario, only 14\% of the size of training dataset is needed to train a more accuracy acoustic model for specific scenario. Our future work includes 1) finding more effective methods to reduce the CER by increasing the number of layers; 2) applying this training framework to Connectionist Temporal Classification (CTC) and attention-based neural networks.

\bibliographystyle{IEEEbib}
\bibliography{deep_lstm}

\begin{thebibliography}{10}

\bibitem{he2015deep}
Kaiming He, Xiangyu Zhang, Shaoqing Ren, and Jian Sun,
\newblock ``Deep residual learning for image recognition,''
\newblock {\em arXiv preprint arXiv:1512.03385}, 2015.

\bibitem{hinton2012deep}
Geoffrey Hinton, Li~Deng, Dong Yu, George~E Dahl, Abdel-rahman Mohamed, Navdeep
  Jaitly, Andrew Senior, Vincent Vanhoucke, Patrick Nguyen, Tara~N Sainath,
  et~al.,
\newblock ``Deep neural networks for acoustic modeling in speech recognition:
  The shared views of four research groups,''
\newblock {\em IEEE Signal Processing Magazine}, vol. 29, no. 6, pp. 82--97,
  2012.

\bibitem{graves2013speech}
Alex Graves, Abdel-rahman Mohamed, and Geoffrey Hinton,
\newblock ``Speech recognition with deep recurrent neural networks,''
\newblock in {\em 2013 IEEE international conference on acoustics, speech and
  signal processing}. IEEE, 2013, pp. 6645--6649.

\bibitem{graves2013hybrid}
Alex Graves, Navdeep Jaitly, and Abdel-rahman Mohamed,
\newblock ``Hybrid speech recognition with deep bidirectional lstm,''
\newblock in {\em Automatic Speech Recognition and Understanding (ASRU), 2013
  IEEE Workshop on}. IEEE, 2013, pp. 273--278.

\bibitem{sak2014long}
Hasim Sak, Andrew~W Senior, and Fran{\c{c}}oise Beaufays,
\newblock ``Long short-term memory recurrent neural network architectures for
  large scale acoustic modeling.,''
\newblock in {\em INTERSPEECH}, 2014, pp. 338--342.

\bibitem{alain2016understanding}
Guillaume Alain and Yoshua Bengio,
\newblock ``Understanding intermediate layers using linear classifier probes,''
\newblock {\em arXiv preprint arXiv:1610.01644}, 2016.

\bibitem{hinton2006reducing}
Geoffrey~E Hinton and Ruslan~R Salakhutdinov,
\newblock ``Reducing the dimensionality of data with neural networks,''
\newblock {\em Science}, vol. 313, no. 5786, pp. 504--507, 2006.

\bibitem{glorot2010understanding}
Xavier Glorot and Yoshua Bengio,
\newblock ``Understanding the difficulty of training deep feedforward neural
  networks.,''
\newblock in {\em Aistats}, 2010, vol.~9, pp. 249--256.

\bibitem{he2015delving}
Kaiming He, Xiangyu Zhang, Shaoqing Ren, and Jian Sun,
\newblock ``Delving deep into rectifiers: Surpassing human-level performance on
  imagenet classification,''
\newblock in {\em Proceedings of the IEEE international conference on computer
  vision}, 2015, pp. 1026--1034.

\bibitem{dean2012large}
Jeffrey Dean, Greg Corrado, Rajat Monga, Kai Chen, Matthieu Devin, Mark Mao,
  Andrew Senior, Paul Tucker, Ke~Yang, Quoc~V Le, et~al.,
\newblock ``Large scale distributed deep networks,''
\newblock in {\em Advances in neural information processing systems}, 2012, pp.
  1223--1231.

\bibitem{zhang2013asynchronous}
Shanshan Zhang, Ce~Zhang, Zhao You, Rong Zheng, and Bo~Xu,
\newblock ``Asynchronous stochastic gradient descent for dnn training,''
\newblock in {\em 2013 IEEE International Conference on Acoustics, Speech and
  Signal Processing}. IEEE, 2013, pp. 6660--6663.

\bibitem{chen2016revisiting}
Jianmin Chen, Rajat Monga, Samy Bengio, and Rafal Jozefowicz,
\newblock ``Revisiting distributed synchronous sgd,''
\newblock {\em arXiv preprint arXiv:1604.00981}, 2016.

\bibitem{mcdonald2010distributed}
Ryan McDonald, Keith Hall, and Gideon Mann,
\newblock ``Distributed training strategies for the structured perceptron,''
\newblock in {\em Human Language Technologies: The 2010 Annual Conference of
  the North American Chapter of the Association for Computational Linguistics}.
  Association for Computational Linguistics, 2010, pp. 456--464.

\bibitem{zinkevich2010parallelized}
Martin Zinkevich, Markus Weimer, Lihong Li, and Alex~J Smola,
\newblock ``Parallelized stochastic gradient descent,''
\newblock in {\em Advances in neural information processing systems}, 2010, pp.
  2595--2603.

\bibitem{chen2016scalable}
Kai Chen and Qiang Huo,
\newblock ``Scalable training of deep learning machines by incremental block
  training with intra-block parallel optimization and blockwise model-update
  filtering,''
\newblock in {\em 2016 IEEE International Conference on Acoustics, Speech and
  Signal Processing (ICASSP)}. IEEE, 2016, pp. 5880--5884.

\bibitem{polyak1992acceleration}
Boris~T Polyak and Anatoli~B Juditsky,
\newblock ``Acceleration of stochastic approximation by averaging,''
\newblock {\em SIAM Journal on Control and Optimization}, vol. 30, no. 4, pp.
  838--855, 1992.

\bibitem{xu2011towards}
Wei Xu,
\newblock ``Towards optimal one pass large scale learning with averaged
  stochastic gradient descent,''
\newblock {\em arXiv preprint arXiv:1107.2490}, 2011.

\bibitem{tian2017}
Tian Xu, Zhang Jun, Ma~Zejun, He~Yi, and Wei Juan,
\newblock ``Exponential moving average model in parallel speech recognition
  training,''
\newblock {\em arXiv preprint arXiv:1703.01024}, 2017.

\bibitem{sak2015fast}
Ha{\c{s}}im Sak, Andrew Senior, Kanishka Rao, and Fran{\c{c}}oise Beaufays,
\newblock ``Fast and accurate recurrent neural network acoustic models for
  speech recognition,''
\newblock {\em arXiv preprint arXiv:1507.06947}, 2015.

\bibitem{hinton2015distilling}
Geoffrey Hinton, Oriol Vinyals, and Jeff Dean,
\newblock ``Distilling the knowledge in a neural network,''
\newblock {\em arXiv preprint arXiv:1503.02531}, 2015.

\bibitem{hinton2006fast}
Geoffrey~E Hinton, Simon Osindero, and Yee-Whye Teh,
\newblock ``A fast learning algorithm for deep belief nets,''
\newblock {\em Neural computation}, vol. 18, no. 7, pp. 1527--1554, 2006.

\bibitem{kingsbury2009lattice}
Brian Kingsbury,
\newblock ``Lattice-based optimization of sequence classification criteria for
  neural-network acoustic modeling,''
\newblock in {\em Acoustics, Speech and Signal Processing, 2009. ICASSP 2009.
  IEEE International Conference on}. IEEE, 2009, pp. 3761--3764.

\bibitem{sak2015learning}
Ha{\c{s}}im Sak, Andrew Senior, Kanishka Rao, Ozan Irsoy, Alex Graves,
  Fran{\c{c}}oise Beaufays, and Johan Schalkwyk,
\newblock ``Learning acoustic frame labeling for speech recognition with
  recurrent neural networks,''
\newblock in {\em Acoustics, Speech and Signal Processing (ICASSP), 2015 IEEE
  International Conference on}. IEEE, 2015, pp. 4280--4284.

\bibitem{sak2015acoustic}
Ha{\c{s}}im Sak, F{\'e}lix de~Chaumont~Quitry, Tara Sainath, Kanishka Rao,
  et~al.,
\newblock ``Acoustic modelling with cd-ctc-smbr lstm rnns,''
\newblock in {\em Automatic Speech Recognition and Understanding (ASRU), 2015
  IEEE Workshop on}. IEEE, 2015, pp. 604--609.

\bibitem{normandin1991hidden}
Yves Normandin,
\newblock {\em Hidden Markov models, maximum mutual information estimation, and
  the speech recognition problem},
\newblock Ph.D. thesis, McGill University, Montreal, 1991.

\bibitem{povey2005discriminative}
Daniel Povey,
\newblock {\em Discriminative training for large vocabulary speech
  recognition},
\newblock Ph.D. thesis, University of Cambridge, 2005.

\bibitem{hermans2013training}
Michiel Hermans and Benjamin Schrauwen,
\newblock ``Training and analysing deep recurrent neural networks,''
\newblock in {\em Advances in Neural Information Processing Systems}, 2013, pp.
  190--198.

\end{thebibliography}

\end{document}